# Power-law Scaling to Assist with Key Challenges in Artificial Intelligence

Yuval Meir[1,*], Shira Sardi[1,*], Shiri Hodassman[1,*], Karin Kisos[1], Itamar Ben-Noam[1], Amir Goldental[1] & Ido Kanter[1,2,†]

[1]Department of Physics, Bar-Ilan University, Ramat-Gan, 52900, Israel.
[2]Gonda Interdisciplinary Brain Research Center, Bar-Ilan University, Ramat-Gan, 52900, Israel.

[*] These authors contributed equally to this work
[†]Corresponding author: ido.kanter@biu.ac.il (I.K.)

**Power-law scaling, a central concept in critical phenomena, is found to be useful in deep learning, where optimized test errors on handwritten digit examples converge as a power-law to zero with database size. For rapid decision making with one training epoch, each example is presented only once to the trained network, the power-law exponent increased with the number of hidden layers. For the largest dataset, the obtained test error was estimated to be in the proximity of state-of-the-art algorithms for large epoch numbers. Power-law scaling assists with key challenges found in current artificial intelligence applications and facilitates an a priori dataset size estimation to achieve a desired test accuracy. It establishes a benchmark for measuring training complexity and a quantitative hierarchy of machine learning tasks and algorithms.**

## Introduction

Phase transition and critical phenomena have been the central focus of statistical mechanics, since the beginning of the second half of 20th century. The thermodynamic properties near the critical point of second-order phase transitions were explained using power-law scaling and hyperscaling relations, depending on the dimensionality of the system[1,2]. The concept of power-law implies a linear relationship between the logarithms of two quantities, that is, a straight line on a log–log plot. It arises from diverse phenomena including the timing and magnitude of earthquakes[3], internet topology and social networks[4-6], turbulence[7], stock price fluctuations[8], word frequencies in linguistics[9] and signal amplitudes in brain activity[10].

Deep learning algorithms are found to be useful in an ever-increasing number of applications, including the analysis of experimental data in physics, ranging from classification problems in astrophysics[11] and high-energy physics data analysis[12] to imaging in noise optics[13] and learning properties of phase transitions[14]. This work indicates that deep learning algorithms behave asymptotically similar to critical physical systems. A basic task in deep learning is supervised learning, where a multilayer network (e.g. Fig. 1a) learns to produce the correct output labels to the input data based on a training database of examples, input–output pairs. A simple example of this is the large Modified National Institute of Standards and Technology (MNIST) database consisting of 60,000 training handwritten digits and 10,000 test digits[15], without any data extension[16,17]. The weights of the selected feedforward network are adjusted using back-propagation algorithm, which is a gradient-descent-based algorithm, to minimize the cost function, thereby, quantifying the mismatch between the current and desired outputs[15].

The performance of the algorithm is estimated using test error, measured on a dataset that was not observed during the training. The test error is expected to decrease with increasing information and increasing dataset size, and to vanish asymptotically in a sufficiently complex network, e.g. enough number of weights, hidden layers and units. The disappearance of the test error with a power-law scaling is the focus of our study, which sets a priori estimation of the required dataset size to achieve the desired test accuracies. The robustness of the power-law scaling phenomenon is examined for training with one and many epochs, that is, for the number of times each example is presented to the trained network, as well as for several feedforward network architectures consisting of a few hidden layers and hyper-weights[18], that is, input crosses. The result of the optimized test errors with one training epoch is in the proximity of state-of-the-art algorithms consisting of a large number of epochs, which has an important implication on the rapid decision making under limited numbers of examples[19,20], which is representative of many aspects of human activity, robotic control[21], and network optimization[22]. The current applicability of the asymptotic test accuracy to such realities using an extremely large number of epochs is questionable. This large gap between advanced learning algorithms and their real-time implementation can be addressed by achieving optimal performance based on only one epoch. Finally, the comparison of the power-law scaling, exponents and constant factors, stem from various learning tasks, datasets, and algorithms is expected to establish a benchmark for a quantitative theoretical framework to measure their complexity[23].

The first trained network that is employed comprises 784 inputs representing 28×28 pixels of a handwritten digit in the range [0, 255] with additional 10,000 input crosses per hidden unit (see Supplementary Appendix A), two hidden layers comprising 100 units each, and 10 outputs representing the labels (Fig. 1a). The presented dataset of examples for the algorithm involves the following initial preprocessing and steps(see Supplementary Appendix A): (a) *Balanced set of examples*: The small dataset consists of an equal number of random examples per label[24]. (b) *Input bias:* The bias of each example is subtracted and the standard deviation of its 784 pixels is normalized to unity. (c) *Fixed order of trained labels*: In each epoch, examples are ordered at random, conditioned to the fixed order of the labels. (d) *Microcanonical set of input crosses:* Each hidden unit in the first layer receives the same number of input crosses, in which each cross comprises two input pixels. (e) *Forward propagation:* A standard sigmoid activation function is attributed to each node[25] and in the forward propagation the accumulative average field is dynamically subtracted from the induced field on each node of the hidden layers.

## Results

### Momentum strategy: Power-law with many epochs

The commonly used learning approach is the backpropagation (BP) strategy given by:

$$W^{t+1} = W^t - \eta \cdot \nabla_{W^t} C \quad (1)$$

where a weight at discrete time-step t, $W^t$, is modified with a step-size η towards the minus sign of the gradient of the cross entropy cost function, C,

$$C = -\frac{1}{M} \sum_{m=1}^{M} [y_m \cdot log(a_m^L) + (1 - y_m) \cdot log(1 - a_m^L)] + \frac{\alpha}{2\eta} \sum_i W_i^2 \quad (2)$$

where $y_m$ stands for the desired labels of the m$^{th}$ examples, $a_m^L$ stands for the current 10 outputs of the output layer L, and the first summation is over all M training examples. The second summation is the overall weights of the network, and $\eta$ and $\alpha$ are constants defined in eqs. (1) and (3), respectively. Here we used the momentum strategy[26]:

$$V^{t+1} = \mu \cdot V^t - \eta \cdot \nabla_{W^t} C$$

$$W^{t+1} = (1 - \alpha) \cdot W^t + V^{t+1} \quad (3)$$

where the friction, μ, and the regularization of the weights, 1-α, are global constants in the region [0, 1] and η is a constant representing the learning rate. In addition there are biases per node associated with the induced field on each node

$$V_b^{t+1} = \mu \cdot V_b^t - \eta \cdot \nabla_{b^t} C$$

$$b^{t+1} = b^t + V_b^{t+1} \quad (4)$$

We minimize the test error for each dataset size over the five parameters of the algorithm $(\mu, \alpha, \eta, Amp_1, Amp_2)$ (where $Amp_i$ are the amplitudes associated with each hidden layer in the forward propagation, see Supplementary Appendix A). The minimized averaged test error, $\epsilon$, for number of examples per label in the range [9,120] indicates a power-law scaling

$$\epsilon \sim \frac{c_0}{(dataset\ size/label)^\rho} \quad (5)$$

with $c_0 \sim 0.65$, $\rho \sim 0.50$ (Fig. 1b), and its extrapolation to the maximal dataset, 6,000 examples per label, indicates a test error of $\epsilon \sim 0.008$. Note that the saturation of the minimal test error is achieved after at least 150 epochs (see Supplementary Appendix B).

## Accelerated strategy: Power-law with many epochs

An accelerated BP method is based on a recent new bridge between experimental neuroscience and advanced artificial intelligence learning algorithms, in which an increased training frequency has been able to significantly accelerate neuronal adaptation processes[24]. This *accelerated* brain-inspired mechanism involves time-dependent step-size, $\eta^t$, associated with each weight, such that coherent consecutive gradients of weight, that is, with the same sign, increase the conjugate $\eta$. The discrete time BP of this accelerated method is summarized for each weight by

$$\eta^{t+1} = \eta^t \cdot e^{-\tau} + A \cdot \tanh(\beta \cdot \nabla_{W^t} C)$$

$$V^{t+1} = \mu \cdot V^t - |\eta^{t+1}| \cdot \nabla_{W^t} C$$

$$W^{t+1} = (1 - \alpha) \cdot W^t + V^{t+1} \qquad (6)$$

where A and β are constants, different for each layer, representing the amplitude and gain, respectively. In addition, there are biases per node similar to eq. (4) where $\eta_0$ is replaced by time-dependent $\eta_b^t$

$$\eta_b^{t+1} = \eta_b^t \cdot e^{-\tau} + A \cdot \tanh(\beta \cdot \nabla_{b^t} C)$$

$$V_b^{t+1} = \mu \cdot V_b^t - |\eta_b^{t+1}| \cdot \nabla_{b^t} C$$

$$b^{t+1} = b^t + V_b^{t+1} \qquad (7)$$

The minimization of the test error of this accelerated method over its 11 parameters $(A_1, A_2, A_3, \beta_1, \beta_2, \beta_3, \mu, \alpha, \tau, Amp_1, Amp_2)$ (see Supplementary Appendix A) is a computational heavy task. It results in the same saturated test error as that for the momentum strategy (Fig. 1b), however, with only 30–50 epochs owing to its accelerated nature.

The test error is further minimized using a soft committee decision based on several replicas, $Nc$, of the network, which are trained on the same set of examples but with different initial weights. The result label, j, for the test accuracy is given by

$$\max_{j}\left(\sum_{s=1}^{N_c} a_{j,s}^{L}\right) \quad (8)$$

where $a_{j,s}^{L}$ stands for the value of the output label j in output layer L and in replica s (j=0, 1, ....9). The minimized test error of the soft committee of the momentum strategy is $\epsilon \sim 0.007$ with $\rho \sim 0.52$ (Fig. 1c), which is in close agreement with state-of-the-art achievements obtained using deep neural networks[27].

## Power-law with one epoch

A similar minimization of the test error, $\epsilon$, is repeated for one epoch, where each example in the training set is presented only once as an input to the feedforward network (Fig. 1a). For the momentum strategy it is found that $\rho \sim 0.49$ and its extrapolation to the maximal dataset (i.e., 6,000 examples per label) results in $\epsilon \sim 0.021$ (Fig. 2a), and for the brain-inspired accelerated strategy in $\epsilon \sim 0.017$ and $\rho \sim 0.49$ (Fig. 2b). For the soft committee of the momentum strategy it is found that $\epsilon \sim 0.015$ with slope, $\rho \sim 0.48$ (Fig. 2a). The test error is reduced even further using soft committee of the accelerated strategy, where $\epsilon \sim 0.013$ with slope, $\rho \sim 0.49$ for 6,000 examples per label (Fig. 2b). Results of one epoch are in the proximity of the test error using many epochs, where the best test error for many epochs $\epsilon \sim 0.007$ has to be compared with $\epsilon \sim 0.013$ for one epoch. These results strongly indicate that rapid decision making, which is representative of many aspects of human activity, robotic control[28], and network optimization[22], is feasible.

## Power-law with several hidden layers

The robustness of the power-law phenomenon for the test error as a function of dataset size (Figs. 1 and 2) is examined for similar feedforward networks without input crosses, and with up to three hidden layers with 100 hidden units each (Fig. 3a). For one hidden layer, the minimization of $\epsilon$ for one epoch and for the momentum strategy indicates $\rho \sim 0.3$, and its extrapolation to 6,000 examples per label results in $\epsilon = 0.053$ (Fig.3b). Using two layers the exponent increases to $\rho \sim 0.34$ with $\epsilon = 0.049$ (Fig. 3c), and for three layers to $\rho = 0.385$ with $\epsilon = 0.048$ (Fig. 3d). These results confirm the existence of the power-law phenomenon in a larger class of feedforward networks and different learning rules as well as the possible increase of the power-law exponent with increasing number of hidden layers (Fig. 3b-d). Asymptotically for very large datasets, increasing the number of hidden layers is expected to minimize $\epsilon$, since $\rho$ increases. However, for a limited number of examples, one layer minimizes $\epsilon$ (Fig. 3b–d), as the constant $c_0$ in eq. (5) is smaller for one layer. Particularly, the power-law scaling indicates that the crossing of $\epsilon$ between one and two layers occurs at ~480 examples per label, whereas the crossing between two and three layers occurs at ~4100 examples per label. This trend stems from the limit of small training datasets and one training epoch, which prevents enhanced optimization of the many more weights of networks with more hidden layers. The asymptotic test error, $\epsilon = 0.049$, of a network with two hidden layers (Fig. 3c) has to be compared with $\epsilon \sim 0.021$ which is achieved for the same architecture with additional input crosses (Fig. 2a). The significant improvement of ~0.028 is attributed to the additional input crosses. This gap also remains under soft committee decision where for two layers without input crosses and the maximal dataset, 6,000 examples per label, $\epsilon \sim 0.039$ (Fig. 4a), which is much greater than $\epsilon \sim 0.007$ (Fig. 1c). We note that $\rho \sim 0.31$ (Fig. 4a) is expected to slightly increase beyond $\rho \sim 0.34$ (Fig. 3c) using better statistics.

## Discussion

The power-law scaling enables the building of an initial step for theoretical framework for deep learning by feedforward neural networks. A classification task, which is characterized by a much smaller power-law exponent, $\rho$, is categorized as a much harder classification problem. It demands a much larger dataset size to achieve the same test error, as long as the constant $c_0$ (eq. (5)) is similar. Similarly, one can compare the efficiency of optimal learning strategy by two different architectures for the same dataset and number of epochs (Figs. 2 and 3) or a comparison of two different BP strategies for the same architecture (Fig. 1). Our work calls for the extension and the confirmation of the power-law scaling phenomenon in other datasets[23,29-32], which will enable to build a hierarchy among their learning complexities. It is especially interesting to observe whether the power-law scaling will lead to a test error in the proximity of state-of-the-art algorithms for other classification and decision problems as well.

The observation in which the test error with one training epoch is in the proximity of the minimized test error using a very large number of epochs paves way for the realization of deep learning algorithms in real-time environments, such as tasks in robotics and network control. A relatively small test error, for instance less than 0.1, can be achieved for a small datasets consisting of only a few tens of examples per label only.

Finally, under the momentum strategy and many training epochs, the minimal saturated test errors of one, two, and three hidden layers and without input crosses are found to be very similar (Fig. 4b). The test error, $\epsilon \sim 0.017$, at the maximal dataset size and $\rho \sim 0.4$ has to be compared to 0.008 with additional input crosses and $\rho \sim 0.5$ (Fig. 1b). For three layers, $\epsilon$ is slightly greater than for one or two layers, but within the error bars. This gap diminishes when the optimized test error for the three layers is obtained under an increased number of epochs, and through the construction of weighs one can show that $\epsilon$ of two layers is achievable with three layers (see Supplementary Appendix F). Furthermore, the similarity of $\epsilon$, independent of the number of hidden layers and for many training epochs (Fig. 4b), is supported by our preliminary results, wherein the average $\epsilon$ of one hidden layer with input crosses and many training epochs is comparable with the one obtained with two hidden layers (Fig. 1b). These results may question the advantage of deep learning based on many hidden layers in comparison to shallow architectures. It is possible that this similarity in the test errors, independent of the number of hidden layers, is either an exceptional case or a larger number of hidden layers enables an easier search in the BP parameters space, which achieves proximity solutions of the minimal test error. However, for the same examined architectures and for one epoch only, the test error and the exponent of the power-law are strongly dependent on the number of hidden layers (Fig. 3).

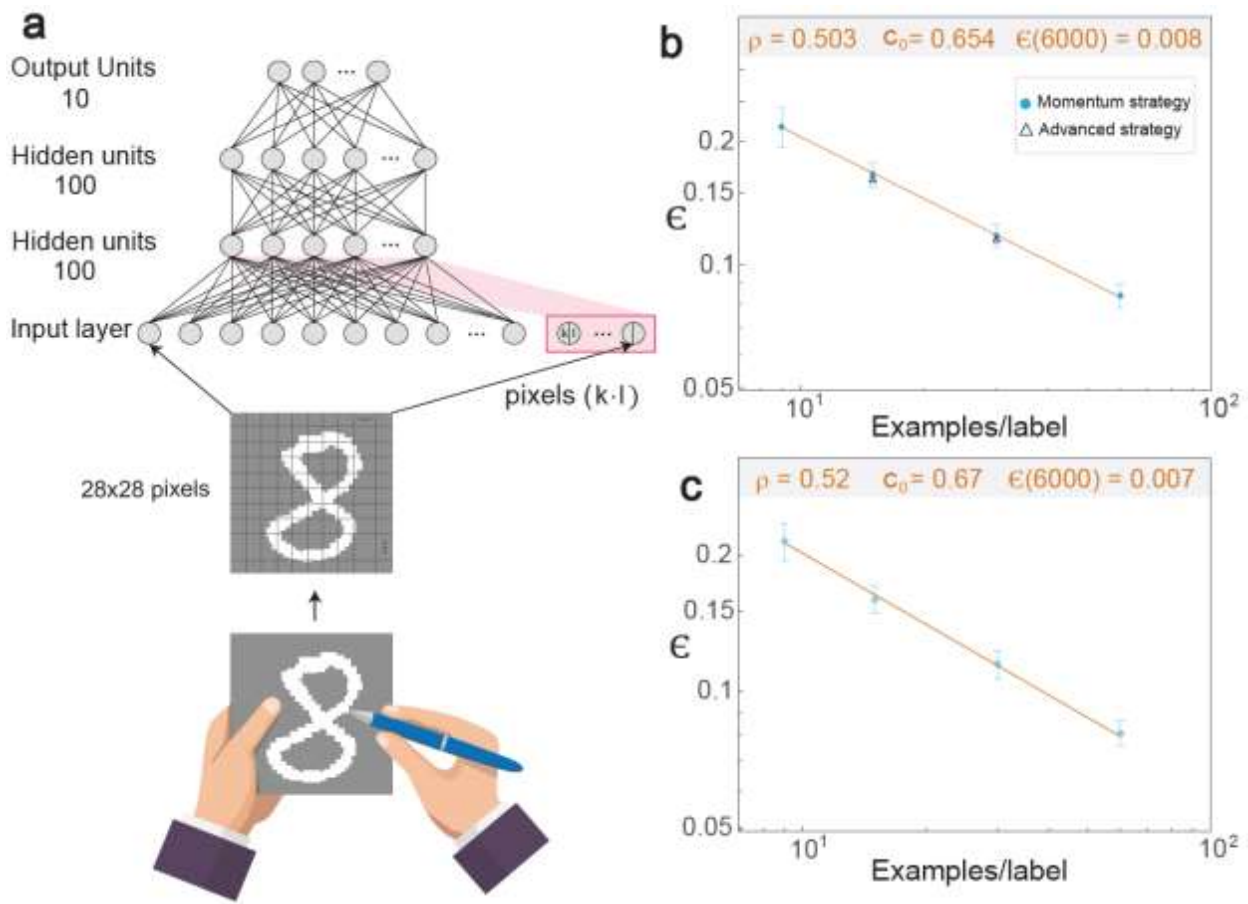

**Figure 1.** Power-law scaling for the test error with many epochs. (a) Scheme of MNIST handwritten digit, which is digitized and fed into the trained network including input crosses (red background). (b) Optimized test error, $\epsilon$, using the architecture in (a), for limited datasets comprising 9, 15, 30 and 60 examples/label and their standard deviations obtained from 50 samples. Momentum strategy (light-blue circles) and acceleration strategy (black triangles). (c) Test error for soft committee decision with $N_c = 50$ (eq. 8). (For details of the parameters, see Supplementary Appendix B)

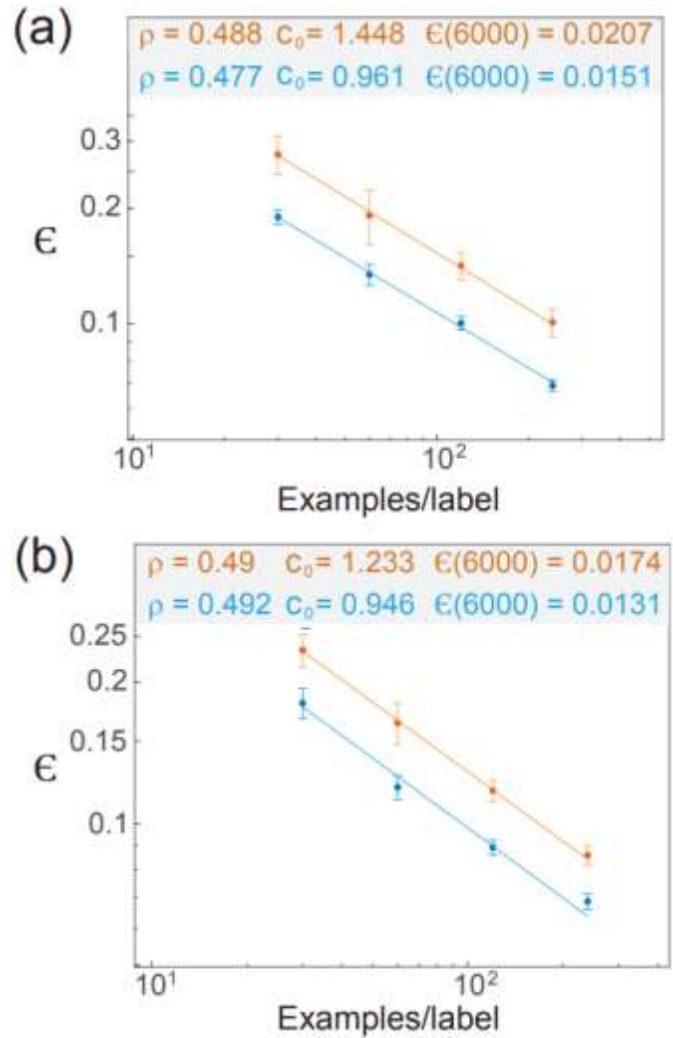

**Figure 2.** Power-law scaling for the test error with one epoch. (a) Test error and its standard deviation as a function of number of examples per label for one epoch only where the trained network is the same as in Fig. 1a. Results for the momentum strategy (orange) and for the soft committee, $N_c = 50$, (blue), where each point is averaged over at least 100 samples. (b) Similar to (a) using the accelerated BP strategy, eqs (6) and (7). (For details of the parameters, see Supplementary Appendix C)

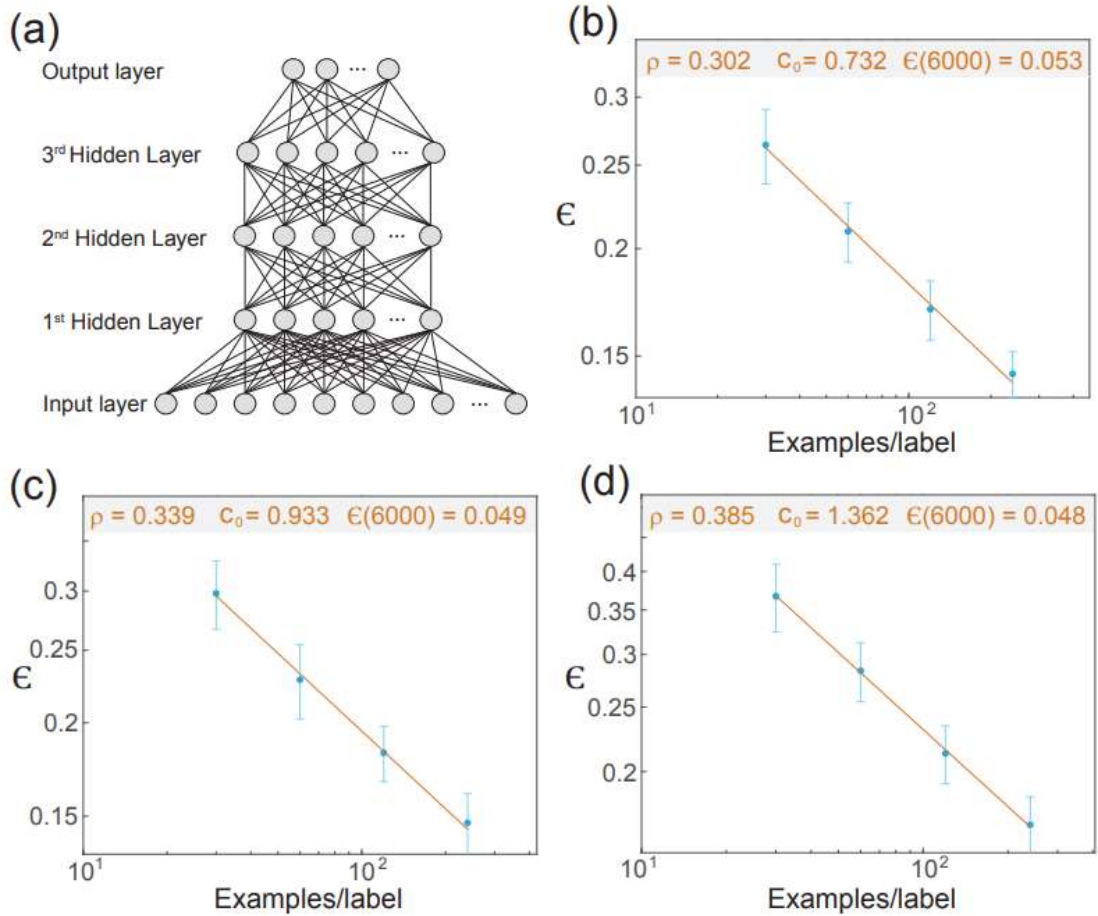

**Figure 3.** Power-law scaling for the test error with several hidden layers and one epoch. (a) Scheme of the trained network on the MNIST examples consisting of three hidden layers having each 100 units and an output layer. In the case of one/two hidden layers only, two/one hidden layers are removed. (b) Minimized test error for 30, 60, 120, and 240 examples/label for one hidden layer (a) using the momentum strategy and one epoch only. The average of each point and its standard deviation are obtained from at least 100 samples. (c) Similar to (b) with two hidden layers in (a). (d) Similar to (b) with three hidden layers in (a). (For details of the parameters, see Supplementary Appendix D)

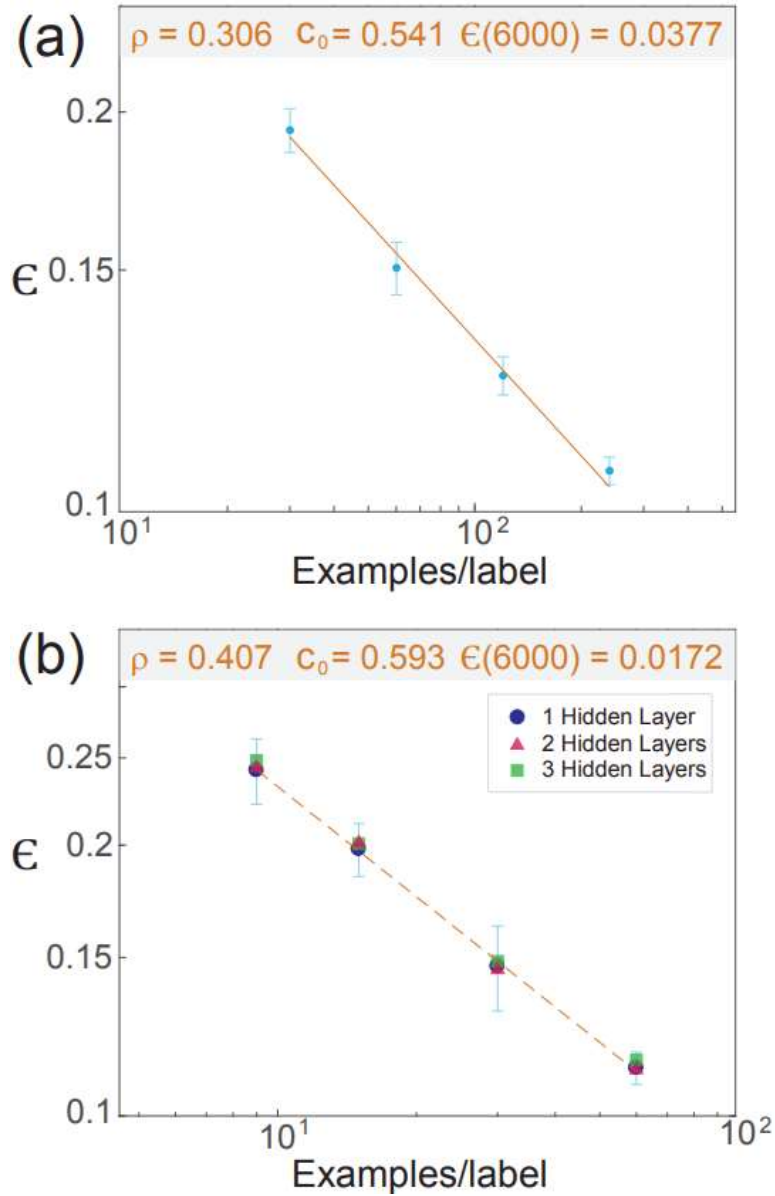

**Figure** 4. (a) Test error, $\varepsilon$, as a function of the number of examples per label for soft committee decision ($N_c$=50 in eq. 8), for two hidden layers without input crosses and one epoch, presented in Fig. 3c. (b) Saturated test error obtained for many epochs as a function of the number of examples per label, for the feedforward network (Fig. 3a), one hidden layer (light-blue circles), two hidden units (orange triangles), and three hidden units (green squares). Typical error bars obtained from at least 200 samples for each number of examples per labels are presented. (For details of the parameters, see Supplementary Appendix E).

**Author contributions**

Y. M., S. S., and S. H. have contributed equally to this work.

**Competing interests**

The authors declare no competing interests.

**Additional information**

**Correspondence** and requests for materials should be addressed to I.K.

# Supplementary Information

**Power-law Scaling to Assist with Key Challenges in Artificial Intelligence**

Yuval Meir[1,*], Shira Sardi[1,*], Shiri Hodassman[1,*], Karin Kisos[1], Itamar Ben-Noam[1], Amir Goldental[1] & Ido Kanter[1,2,†]

[1]Department of Physics, Bar-Ilan University, Ramat-Gan, 52900, Israel.
[2]Gonda Interdisciplinary Brain Research Center, Bar-Ilan University, Ramat-Gan, 52900, Israel.

[*] These authors contributed equally to this work
[†]Corresponding author: ido.kanter@biu.ac.il (I.K.)

## APPENDIX A: DETAILS OF THE USED ALGORITHM

**Architecture and initial weights:** The feedforward neural network (Fig. 1) consists of 784 input units with additional 10,000 input-crosses for each hidden unit (see Input), 2 hidden layers consist of 100 units each and 10 output units. Weights between successive layers are fully connected, except the input-crosses. Each unit in the hidden and the output layers has an additional input from a bias unit. We denote by $W^1$, $W^2$ and $W^3$ the weights from the input layer to the first hidden layer, from the first hidden layer to the second hidden one and from the second hidden layer to the output layer, respectively. The initial conditions of all weights are randomly chosen from a Gaussian distribution with a zero average and standard deviation (STD) equals 1. All weights are normalized such that all input weights to each hidden unit have a zero average and an STD equals 1[1].

After the above-mentioned initial normalization of all weights, the weights of the input-crosses are rescaled

$$W_{\text{input crosses}} = \sqrt{\frac{\#\text{ regular input}}{\#\text{ input crosses}}} \cdot W_{\text{input crosses}} = \sqrt{\frac{784}{10000}} \cdot W_{\text{input crosses}}$$

such that initially they have the same effect on the forward propagation as the regular weights. In addition, the initial value of the bias of each weights is set to 1.

**Input:** Each example, $\widetilde{X}_m$, $m = 1, 2, \ldots, M$, of the train dataset consists of 784 pixels, $\widetilde{X}_{m,p}$, which their values are in the range [0, 255]. The input, X, of the example $\widetilde{X}$, consists of the original 784 pixels where the average pixel value in $\widetilde{X}$ is subtracted from each pixel and the standard deviation is set to one:

$$X_m = \widetilde{X}_m - \frac{1}{784}\sum_{p=1}^{784}\widetilde{X}_{m,p}$$
$$X_m = X_m/\text{std}(\widetilde{X}_m)$$

Furthermore, an input pixel which has an identical value among all the training examples, e.g. have zero variance in all train dataset examples, is set to zero.

An addition of 10,000 input-crosses are added to the input, $X_{k,l}$:

$$X_{k,l} = X_k \cdot X_l$$

where k and l are random indices in the range [1, 784] with corresponding pixels $X_k$ and $X_l$ for a given example. We forbid input-crosses that are zero in all the train dataset. Each input-cross is not connected more than once to each hidden unit.

**Forward propagation:** The output of a unit, j, in the first hidden layer for the $m^{th}$ example, for instance, $a^1_{j,m}$, is calculated as:

$$z^1_{j,m} = \sum_j (W^1_{ij} \cdot X_i) + b^1_j$$

$$z^1_{j,m} = z^1_{j,m} - \text{Amp}_1 \cdot \frac{1}{m-1} \sum_{t=1}^{m-1} z^1_{j,t}$$

$$a^1_{j,m} = \frac{1}{1 + e^{-z^1_{j,m}}}$$

where $W^1_{ij}$ is the weight from the $i^{th}$ input unit to the $j^{th}$ hidden unit, $X_i$ is the $i^{th}$ input and $b^1_j$ is the bias induced on the $j^{th}$ unit in the first hidden layer. $z^1_{j,m}$ represents the field propagating from the input layer. Each time we calculate the field, $z^1_{j,m}$, we subtract the accumulative average field for the input layer of the previous m-1 examples, where $\text{Amp}_1$ is a constant representing the amplitude of reduction. Note that $z^1_{j,m}$ is not modified for m = 1.

For the second hidden layer, the output of the $j^{th}$ unit for the $m^{th}$ example, $a^2_{j,m}$, is calculated as following:

$$z^2_{j,m} = \sum_j (W^2_{ij} \cdot a^1_{j,m}) + b^2_j$$

$$z^2_{j,m} = z^2_{j,m} - \text{Amp}_2 \cdot \frac{1}{m-1} \sum_{t=1}^{m-1} z^2_{j,t}$$

$$a^2_{j,m} = \frac{1}{1 + e^{-z^2_{j,m}}}$$

where $W_{ij}^2$ is the weight from the $i^{th}$ unit in the first hidden layer to the $j^{th}$ unit in the second hidden layer, and $b_j^2$ is the bias induced on the $j^{th}$ unit in the second hidden layer. $z_{j,m}^2$ represents the field for the second layer. Each time we calculate the field, $z_{j,m}^2$, we subtract the accumulative average field for the second layer of the previous m-1 examples, where $Amp_2$ is a constant representing the amplitude of reduction. Note that $z_{j,m}^2$ is not modified for m = 1.

The output of the $j^{th}$ unit in the output layer, $a_j^3$, is calculated as following:

$$z_{j,m}^3 = \sum_j (W_{ij}^3 \cdot a_{j,m}^2) + b_j^3$$

$$a_{j,m}^3 = \frac{1}{1 + e^{-z_{j,m}^3}}$$

where $W_{ij}^3$ is the weight from the $i^{th}$ unit in the second hidden layer to the $j^{th}$ output unit, and $b_j^3$ is the bias induced on the $j^{th}$ output unit.

**Back propagation:** We use the cross entropy cost function

$$C = -\frac{1}{M} \sum_{m=1}^{M} [y_m \cdot \log(a_m) + (1-y_m) \cdot \log(1-a_m)] + \frac{\alpha}{2\eta} \sum_i W_i^2$$

where $y_m$ stands for the desired labels and $a_m$ stands for the current 10 output units of the output layer and $\eta$ and $\alpha$ are constants defined in eqs. (1) and (3) in the main text, respectively. The summation is over all M training examples. The second summation is over all weights of the network. Note that for the accelerated strategy, $\eta = \eta^t$ in the above cost function.

The backpropagation method computes the gradient for each weight with respect to the cost function. The weights and biases are updated according to the advanced acceleration method[1]:

$$\eta^{t+1} = \eta^t \cdot e^{-\tau} + A_{1/2/3} \cdot \tanh(\beta_{1/2/3} \cdot \nabla_{W^t} C)$$
$$V^{t+1} = \mu \cdot V^t - |\eta^{t+1}| \cdot \nabla_{W^t} C$$

$$W^{t+1} = (1-\alpha) \cdot W^t + V^{t+1}$$

$$\eta_b^{t+1} = \eta_b^t \cdot e^{-\tau} + A_d \cdot \tanh(\beta_d \cdot \nabla_{b^t} C)$$
$$V_b^{t+1} = \mu \cdot V_b^t - |\eta_b^{t+1}| \cdot \nabla_{b^t} C$$
$$b^{t+1} = b^t + V_b^{t+1}$$

where t is the discrete time-step, W are the weights, 1-$\alpha$ is a regularization constant and $\eta$ is defined for each weight. $A_d$ and $\beta_d$ are constants representing the amplitude and the gain between the $d^{th}$ and $d+1^{th}$ layers, d=1,2 and 3. $\eta$ is initialized as: $\eta_0 = A_d \cdot \tanh(\beta_d \cdot \nabla_W C_{first})$, where $\nabla_W C_{first}$ is the first computed gradient. V is initialized as: $V_0 = -|\eta_0| \cdot \nabla_W C_{first}$.

**Test accuracy:** The network test accuracy is calculated based on the MNIST dataset for testing, containing 10,000 input examples. The test examples are modified in the same way as the examples in the training dataset.

**Optimization:** The selection of the optimized parameters. For a given architecture and number of epochs, the optimization procedure first evaluates the test error over a rough grid of the adjustable parameters followed by fine-tuning grids with higher resolutions. For example, the $\alpha$ parameter in the range (0, 1) was first estimated under a rough grid $\Delta\alpha = 0.1$. Next, the selected range for further optimization (0, 0.1), for instance, was estimated under a resolution $\Delta\alpha = 0.01$, and finally under a resolution of $\Delta\alpha = 0.0001$. The maximal resolution was selected such that the test error for a desired resolution was unaffected by selecting a higher resolution. All other tunable parameters were optimized similarly. Note that the training error practically vanishes. For the momentum strategy and small dataset sizes, a search over the entire selected grid was possible. However, for large dataset sizes and for the acceleration strategy consists of 11 parameters an optimization of the test accuracy over a grid was beyond our computational capabilities. We note, that in order to obtain a meaningful optimization procedure, we need to average each measured point over 20-50 different samples, otherwise, the optimization procedure is dominated by stochastic fluctuations.

In cases where a complete optimization over a grid was impossible, we optimized sequentially each parameter over its grid. Nevertheless, we confirmed that a few different sequential orders of the optimized parameters result in the same optimized test accuracy and set of parameters.

The optimization is performed independently for each examined dataset size, number of examples and number of epochs. Results for the committee systems are based on the optimized selected parameters for a single system. The optimized parameters are summarized in the following tables.

We note that cross validation was confirmed using several validation databases consisting each of 10,000 random examples with the same statistics for each label as in the test set. Averaged results were in the same STD of the reported test errors. In addition, preliminary results also indicate that databases consisting of random selected examples, with different fluctuations for each label, also result in similar test errors.

## APPENDIX B: FIGURE 1 – OPTIMIZED PARAMTERS

**Figure 1b optimized parameters:**

| Momentum strategy - Parameters | | | | | | |
|---|---|---|---|---|---|---|
| Examples/label | $\eta$ | $\mu$ | $\alpha$ | $Amp_1$ | $Amp_2$ | Epoch |
| 9 | 0.004 | 0.95 | 0.0001 | 0.06 | 0.03 | 150 |
| 15 | 0.0095 | 0.65 | 0.00022 | 0.08 | 0.04 | 150 |
| 30 | 0.008 | 0.771 | 0.00048 | 0.07 | 0.045 | 150 |
| 60 | 0.0005 | 0.9555 | 0.0003 | 0.1 | 0.006 | 150 |

| Momentum strategy - Classifications | | | |
|---|---|---|---|
| Examples/label | Epoch | Success rate | Std |
| 9 | 150 | 0.783 | ± 0.0244 |
| 15 | 150 | 0.8336 | ± 0.0114 |
| 30 | 150 | 0.8823 | ± 0.0076 |
| 60 | 150 | 0.916 | ± 0.0056 |

| Accelerated strategy - Parameters | | | | | | | | | | | | |
|---|---|---|---|---|---|---|---|---|---|---|---|---|
| Examples/label | $A_1$ | $A_2$ | $A_3$ | $\beta_1$ | $\beta_2$ | $\beta_3$ | $\mu$ | $\alpha$ | $\tau$ | $Amp_1$ | $Amp_2$ | Epoch |
| 15 | 0.04 | 0.04 | 0.004 | 1000 | 1000 | 50 | 0.005 | 0.001 | 0.094 | 0.004 | 0.004 | 50 |
| 30 | 0.04 | 0.04 | 0.004 | 1000 | 1000 | 50 | 0.004 | 0.001 | 0.092 | 0.04 | 0.04 | 30 |

| Accelerated strategy - Classifications | | | |
|---|---|---|---|
| Examples/label | Epoch | Success rate | Std |
| 15 | 50 | 0.8391 | ± 0.0122 |
| 30 | 30 | 0.8854 | ± 0.0066 |

**Figure 1c optimized parameters:**

| Momentum strategy - Classifications | | | | |
|---|---|---|---|---|
| Examples/label | Epoch | $N_c$ Committee | Success rate committees | Std committee |
| 9 | 150 | 101 | 0.8 | ± 0.0203 |
| 15 | 150 | 101 | 0.84 | ± 0.0115 |
| 30 | 150 | 101 | 0.8865 | ± 0.0083 |
| 60 | 150 | 101 | 0.919 | ± 0.0042 |

The parameters used in this figure are the same as in Figure 1b.

## APPENDIX C: FIGURE 2 – OPTIMIZED PARAMTERS

### Figure 2a optimized parameters:

| Momentum strategy – one epoch | | | | | | |
|---|---|---|---|---|---|---|
| Examples/ label | $\eta$ | $\mu$ | $\alpha$ | $Amp_1$ | $Amp_2$ | Epoch |
| 30 | 0.0043 | 0.955 | 0.0065 | 0.2 | 0.004 | 1 |
| 60 | 0.0025 | 0.9555 | 0.004 | 0.19 | 0.004 | 1 |
| 120 | 0.0021 | 0.95 | 0.0016 | 0.2 | 0.04 | 1 |
| 240 | 0.003 | 0.91 | 0.0013 | 0.2 | 0.004 | 1 |

| Momentum strategy – one epoch | | | | | | |
|---|---|---|---|---|---|---|
| Examples/ label | Epoch | $N_c$ Committee | Success rate | Std | Success rate committees | Std committee |
| 30 | 1 | 101 | 0.723 | ± 0.032 | 0.81 | ± 0.0082 |
| 60 | 1 | 101 | 0.808 | ± 0.0312 | 0.8655 | ± 0.0084 |
| 120 | 1 | 101 | 0.8583 | ± 0.0117 | 0.8996 | ± 0.0041 |
| 240 | 1 | 101 | 0.8991 | ± 0.0085 | 0.931 | ± 0.0015 |

### Figure 2b optimized parameters:

| Accelerated strategy – one epoch - Parameters | | | | | | | | | | | | |
|---|---|---|---|---|---|---|---|---|---|---|---|---|
| Examples/ label | $A_1$ | $A_2$ | $A_3$ | $\beta_1$ | $\beta_2$ | $\beta_3$ | $\mu$ | $\alpha$ | $\tau$ | $Amp_1$ | $Amp_2$ | Epoch |
| 30 | 0.005 | 0.015 | 0.00001 | 55 | 35 | 20 | 0.85 | 0.007 | 0.01 | 0.18 | 0.22 | 1 |
| 60 | 0.02 | 0.0405 | 0.00066 | 2490 | 3050 | 450 | 0.61 | 0.0046 | 0.054456 | 0.21 | 0.005 | 1 |
| 120 | 0.017 | 0.00515 | 0.0005 | 5000 | 2000 | 50 | 0.625 | 0.002 | 0.040822 | 0.205 | 0.0013 | 1 |
| 240 | 0.004 | 0.004 | 0.0003 | 1400 | 1800 | 5 | 0.63 | 0.00117 | 0.0095 | 0.005 | 0.002 | 1 |

| Accelerated strategy – one epoch - Classifications | | | | | | |
|---|---|---|---|---|---|---|
| Examples/ label | Epoch | $N_c$ Committee | Success rate | Std | Success rate committees | Std committee |
| 30 | 1 | 101 | 0.766 | ± 0.0189 | 0.815 | ± 0.0138 |
| 60 | 1 | 101 | 0.8361 | ± 0.0165 | 0.881 | ± 0.0119 |
| 120 | 1 | 101 | 0.8815 | ± 0.0021 | 0.911 | ± 0.0033 |
| 240 | 1 | 101 | 0.9142 | ± 0.0044 | 0.932 | ± 0.002 |

**Figure 3b optimized parameters:**

| Momentum strategy - 1 hidden layers | | | | | |
|---|---|---|---|---|---|
| Examples/ label | η | μ | α | $Amp_1$ | Epoch |
| 30 | 0.005 | 0.945 | 0.007 | 0.15 | 1 |
| 60 | 0.0033 | 0.942 | 0.004 | 0.11 | 1 |
| 120 | 0.002 | 0.948 | 0.0021 | 0.13 | 1 |
| 240 | 0.0025 | 0.945 | 0.0008 | 0.003 | 1 |

| Momentum strategy - 1 hidden layers | | | |
|---|---|---|---|
| Examples/ label | Epoch | Success rate | Std |
| 30 | 1 | 0.736 | ± 0.0262 |
| 60 | 1 | 0.7906 | ± 0.0166 |
| 120 | 1 | 0.83 | ± 0.0135 |
| 240 | 1 | 0.857 | ± 0.0088 |

# APPENDIX D: FIGURE 3 – OPTIMIZED PARAMTERS

**Figure 3c optimized parameters:**

| Momentum strategy - 2 hidden layers | | | | | | |
|---|---|---|---|---|---|---|
| Examples/ label | $\eta$ | $\mu$ | $\alpha$ | $Amp_1$ | $Amp_2$ | Epoch |
| 30 | 0.006 | 0.95 | 0.0066 | 0.15 | 0.005 | 1 |
| 60 | 0.0037 | 0.962 | 0.0032 | 0.15 | 0.2 | 1 |
| 120 | 0.0022 | 0.97 | 0.0015 | 0.15 | 0.2 | 1 |
| 240 | 0.002 | 0.97 | 0.00084 | 0.05 | 0.2 | 1 |

| Momentum strategy - 2 hidden layers | | | |
|---|---|---|---|
| Examples/ label | Epoch | Success rate | Std |
| 30 | 1 | 0.7013 | ± 0.0313 |
| 60 | 1 | 0.7717 | ± 0.026 |
| 120 | 1 | 0.8176 | ± 0.0155 |
| 240 | 1 | 0.854 | ± 0.012 |

**Figure 3d optimized parameters:**

| Momentum strategy - 3 hidden layers | | | | | | | |
|---|---|---|---|---|---|---|---|
| Examples/ label | $\eta$ | $\mu$ | $\alpha$ | $Amp_1$ | $Amp_2$ | $Amp_3$ | Epoch |
| 30 | 0.0078 | 0.933 | 0.0046 | 0.34 | 0.14 | 0.13 | 1 |
| 60 | 0.0043 | 0.946 | 0.0023 | 0.25 | 0.017 | 0.0018 | 1 |
| 120 | 0.0046 | 0.943 | 0.001 | 0.47 | 0.12 | 0.9 | 1 |
| 240 | 0.003 | 0.945 | 0.0007 | 0.13 | 0.1 | 0.001 | 1 |

| Momentum strategy - 3 hidden layers | | | |
|---|---|---|---|
| Examples/ label | Epoch | Success rate | Std |
| 30 | 1 | 0.6328 | ± 0.0428 |
| 60 | 1 | 0.7164 | ± 0.0289 |
| 120 | 1 | 0.787 | ± 0.0213 |
| 240 | 1 | 0.8337 | ± 0.0169 |

## APPENDIX E: FIGURE 4 – OPTIMIZED PARAMTERS

**Figure 4a optimized parameters:**

| Momentum strategy - 2 hidden layers | | | | |
|---|---|---|---|---|
| Examples/ label | Epoch | $N_c$ Committee | Success rate committee | Std committee |
| 30 | 1 | 101 | 0.8065 | ± 0.0077 |
| 60 | 1 | 101 | 0.8494 | ± 0.0072 |
| 120 | 1 | 101 | 0.8762 | ± 0.0043 |
| 240 | 1 | 101 | 0.8959 | ± 0.0026 |

The parameters used in this figure are the same as in Figure 3c.

**Figure 4b optimized parameters:**

| Momentum strategy - 1 hidden layers | | | | | |
|---|---|---|---|---|---|
| Examples/ label | $\eta$ | $\mu$ | $\alpha$ | $Amp_1$ | Epoch |
| 9 | 0.0006 | 0.945 | 0.0008 | 0.095 | 300 |
| 15 | 0.0004 | 0.95 | 0.0005 | 0.1 | 300 |
| 30 | 0.0054 | 0.978 | 0.00003 | 0.000095 | 300 |
| 60 | 0.0000089 | 0.9999 | 0.000017 | 0.0000975 | 300 |

| Momentum strategy - 1 hidden layers | | | |
|---|---|---|---|
| Examples/ label | Epoch | Success rate | Std |
| 9 | 300 | 0.7577 | ± 0.0202 |
| 15 | 300 | 0.802 | ± 0.0142 |
| 30 | 300 | 0.8533 | ± 0.0138 |
| 60 | 300 | 0.8869 | ± 0.0047 |

| Momentum strategy - 2 hidden layers | | | | | | |
|---|---|---|---|---|---|---|
| Examples/ label | $\eta$ | $\mu$ | $\alpha$ | $Amp_1$ | $Amp_2$ | Epoch |
| 9 | 0.00075 | 0.999 | 0.0001 | 0.001 | 0.01 | 300 |
| 15 | 0.0005 | 0.85 | 0.00007 | 0.01 | 0.004 | 300 |
| 30 | 0.0054 | 0.978 | 0.00003 | 0.000095 | 0.046 | 300 |
| 60 | 0.0000092 | 0.9999 | 0.000016 | 0.000098 | 0.047 | 300 |

| Momentum strategy - 2 hidden layers | | | |
|---|---|---|---|
| Examples/ label | Epoch | Success rate | Std |
| 9 | 300 | 0.7561 | ± 0.02 |
| 15 | 300 | 0.7988 | ± 0.0137 |
| 30 | 300 | 0.8546 | ± 0.0123 |
| 60 | 300 | 0.8875 | ± 0.0052 |

| Momentum strategy - 3 hidden layers | | | | | | | |
|---|---|---|---|---|---|---|---|
| Examples/ label | $\eta$ | $\mu$ | $\alpha$ | $Amp_1$ | $Amp_2$ | $Amp_3$ | Epoch |
| 9 | 0.0078 | 0.933 | 0.0046 | 0.34 | 0.14 | 0.13 | 1200 |
| 15 | 0.008 | 0.968 | 0.00004 | 0.001 | 0.02 | 0.005 | 500 |
| 30 | 0.008 | 0.968 | 0.00004 | 0.001 | 0.02 | 0.005 | 300 |
| 60 | 0.0065 | 0.971 | 0.00002 | 0.01 | 0.005 | 0.0001 | 600 |

| Momentum strategy - 3 hidden layers | | | |
|---|---|---|---|
| Examples/ label | Epoch | Success rate | Std |
| 9 | 1200 | 0.7515 | ± 0.0181 |
| 15 | 500 | 0.7992 | ± 0.0149 |
| 30 | 300 | 0.8514 | ± 0.0142 |
| 60 | 600 | 0.8847 | ± 0.0084 |

# APPENDIX F: TWO AND THREE HIDDEN LAYERS WITH THE SAME PERFORMANCE

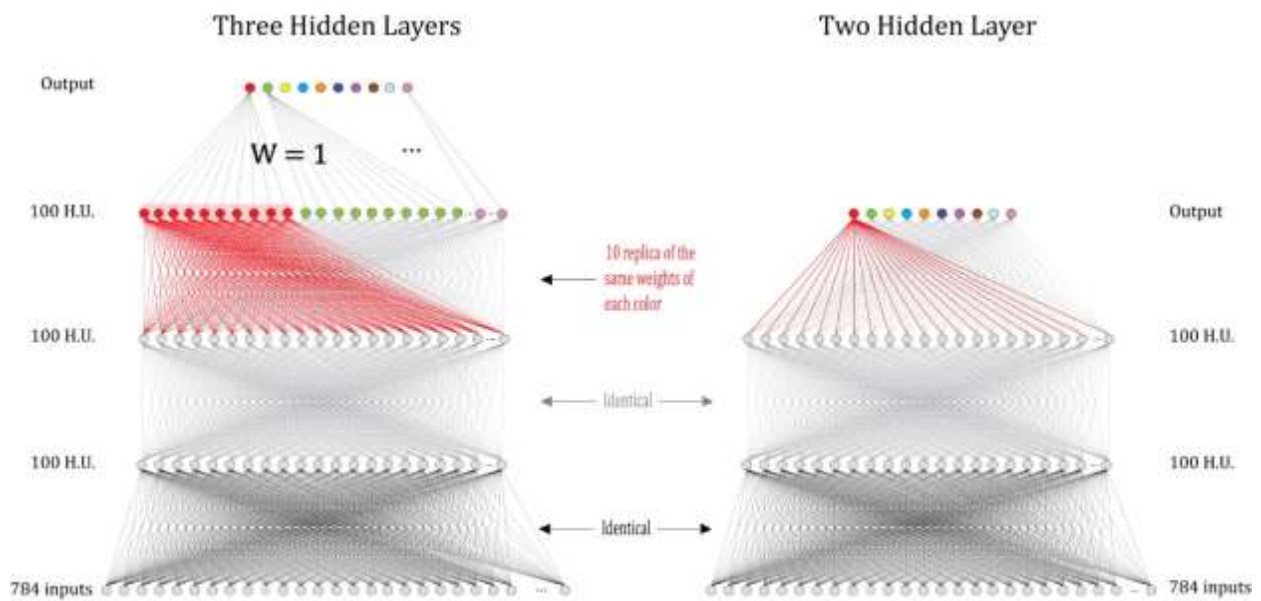

FIG. 5. A special construction of the three hidden layers weights result in the same test error as for the same architecture but with two hidden layers only (Fig. 3a).

An example of three hidden layers neural network consisting of 784 input units, 100 hidden units for each one of the three hidden layers and 10 output units (left), and the same architecture with only two hidden layers (right). The weights between the input layer and the first hidden layer, along with the weights between the first hidden layer and the second hidden layer are identical for both networks. For the third hidden layer (left), every 10 hidden units (e.g. 10 red units) replicate the incoming weights and the output of one of the output unit (e.g. red) in the right architecture. Finally, each group of the 10 hidden units with the same color in the third hidden layer (left) are connected to a distinct output unit using a constant weight, e.g. W=1. The rest of the weights between the third hidden layer and the output layer vanish. Note that the output is different between the two networks because of the nonlinear activation function of the output units. However, the two networks make the same decision for any input, i.e. the maximal label of both networks is the same, since the nonlinear activation function is monotonically increasing.